\documentclass{article}

\usepackage{PRIMEarxiv}
\usepackage{xcolor}

\usepackage[utf8]{inputenc} 
\usepackage[T1]{fontenc}    
\usepackage{hyperref}       
\usepackage{url}            
\usepackage{booktabs}       
\usepackage{amsfonts}       
\usepackage{nicefrac}       
\usepackage{microtype}      
\usepackage{lipsum}
\usepackage{fancyhdr}       
\usepackage{graphicx}       
\graphicspath{{media/}}     
\usepackage{amsmath}
\usepackage{cite}

\usepackage{titlesec}
\titleformat{\paragraph}[block]
  {\normalfont\normalsize\bfseries}   
  {\theparagraph}                     
  {1em}                               
  {}                                  
  []                                  

\titlespacing*{\paragraph}
  {0pt}    
  {1.5ex plus 0.5ex minus .2ex}  
  {0.8ex plus .2ex}              

\title{A Machine Learning–Based Multimodal Framework for Wearable Sensor-Based Archery Action Recognition and Stress Estimation}

\author{
  Xianghe Liu\textsuperscript{†}, Jiajia Liu, Chuxian Xu, Minghan Wang, Hongbo Peng, Tao Sun, Jiaqi Xu\textsuperscript{*} \\
  Beijing PsychTech Technology Co., Ltd.\\
  Beijing, China\\
}

\begin{document}
\maketitle
\noindent\textsuperscript{†} First author. \\
\noindent\textsuperscript{*} Corresponding author (Email: kylexu0626@icloud.com).\par

\begin{abstract}

In precision sports such as archery, athletes’ performance depends on both biomechanical stability and psychological resilience. Traditional motion analysis systems are often expensive and intrusive, limiting their use in natural training environments. To address this limitation, we propose a machine learning–based multimodal framework that integrates wearable sensor data for simultaneous action recognition and stress estimation. Using a self-developed wrist-worn device equipped with an accelerometer and photoplethysmography (PPG) sensor, we collected synchronized motion and physiological data during real archery sessions.
For motion recognition, we introduce a novel feature—Smoothed Differential Acceleration (SmoothDiff)—and employ a Long Short-Term Memory (LSTM) model to identify motion phases, achieving 96.8\% accuracy and 95.9\% F1-score. For stress estimation, we extract heart rate variability (HRV) features from PPG signals and apply a Multi-Layer Perceptron (MLP) classifier, achieving 80\% accuracy in distinguishing high- and low-stress levels.
The proposed framework demonstrates that integrating motion and physiological sensing can provide meaningful insights into athletes’ technical and mental states. This approach offers a foundation for developing intelligent, real-time feedback systems for training optimization in archery and other precision sports.

\end{abstract}


\section{Introduction}

In precision sports such as archery, athletes’ success depends not only on precise and repeatable body movements but also on their ability to manage psychological stress under competitive conditions \cite{kim2018effects, Kayacan2021The}. Even minor deviations in form or increases in anxiety can critically impact shot accuracy and overall performance \cite{Kayacan2021The}. Providing athletes and coaches with integrated insights on both movement patterns and psychological stress is essential for optimizing training and competition outcomes \cite{Mitchell2024Integrating}.

Traditional performance analysis methods in archery—such as laboratory-based motion capture systems or camera setups—offer detailed kinematic analysis but are often costly, intrusive, and impractical for routine use outside controlled environments \cite{Suo2024Digital}. To address these challenges, acceleration sensors have been applied in archery to assess motion phases and tremors in a less intrusive manner \cite{Ogasawara2021Archery}. At the same time, advances in wearable sensing technologies—particularly wrist-worn devices equipped with accelerometers and photoplethysmography (PPG) sensors—have enabled non-invasive, continuous monitoring of both movement and physiological signals, such as heart rate, during dynamic activities like running or fitness training \cite{Mullan2015Unobtrusive, Zhang2015Photoplethysmography-Based}. However, the integration of PPG and accelerometer data has not yet been systematically explored in archery-specific contexts, where fine motor control and psychological stress are critical to performance.

Despite these advances, current research often treats motion analysis and stress assessment as separate tasks, missing the opportunity to study their interactions. Moreover, few studies have focused specifically on archery, and no prior work has developed integrated models using synchronized motion and physiological data collected under naturalistic conditions.

To address these gaps, we present an archery-focused wearable sensing system that combines accelerometer-based motion recognition with PPG-based stress detection. We construct a dataset containing synchronized accelerometer and PPG signals, annotated with expert-defined motion phases (draw, aim, release) and athletes’ self-reported stress ratings. For motion recognition, we propose a novel feature—Smoothed Differential Acceleration (SmoothDiff)—to improve the accuracy of phase detection using Long Short-Term Memory (LSTM) networks. For stress detection, we extract heart rate variability (HRV) features from the PPG signals and train a Multi-Layer Perceptron (MLP) model to classify stress levels.

Our contributions can be summarized as follows:
\begin{itemize}
    \item We develop a specialized dataset combining synchronized motion (accelerometer) and physiological (PPG) data, labeled for motion phases and subjective stress levels, to enable model development and evaluation.
    \item We propose a novel feature extraction method (SmoothDiff) that significantly enhances motion recognition performance.
    \item We develop and validate deep learning models that achieve accurate simultaneous recognition of archery movements and psychological stress, providing a foundation for real-world feedback applications.
\end{itemize}

This work lays the groundwork for wearable-based monitoring systems in archery, offering data-driven insights into athletes’ movement patterns and psychological states to support training and resilience.

\section{Methods\label{Dataset}
}

\subsection{Participants and Apparatus}
This study involved six adolescent professional archers (4 female, 2 male), aged 14.9 to 16.3 years, with a mean age of 15.5 years (SD = 0.5).

We used a self-developed intelligent wristband device (Ergosensing, China) worn on the athletes’ bow hand (left wrist for all participants). The device is equipped with an accelerometer and photoplethysmography (PPG) sensor. The accelerometer recorded three-axis acceleration data (Acc X, Acc Y, Acc Z) at a sampling rate of 20 Hz, while the PPG sensor measured physiological signals using reflected green light at 532 nm wavelength, also at a sampling rate of 20 Hz. This device has been validated and applied in prior physiological monitoring research, demonstrating reliability for non-invasive signal collection \cite{zhang2021cped}.

\subsection{Experimental Procedure}
The experiment was conducted under official archery competition conditions. Each participant completed multiple shooting rounds, with each round consisting of three arrows. The total number of rounds varied depending on the athletes’ performance, as higher-performing athletes advanced to subsequent stages.

During the experiment, a handheld tablet was used by the experimenter to mark key events in real time. Specifically, at the start of each round, an ExpStart marker was recorded, and at the end of each round, an ExpEnd marker was added. Within each round, the experimenter also marked each draw action (Draw) and each release action (Release). These annotations were synchronized with the wristband’s accelerometer and PPG recordings, ensuring precise alignment of motion and physiological signals for subsequent analysis.

At the end of each round, participants were asked to self-report their perceived stress level using a five-point Likert scale, where 1 indicated no stress and 5 indicated high stress. This self-reported stress score was used as the ground-truth label for the stress detection task. All experimental sessions were conducted on the same day, under consistent environmental conditions.

\label{subsec:data gathering}



\subsection{Data Preprocessing}

\subsubsection{Accelerometer Signal Processing}

The raw accelerometer data (Acc X, Acc Y, Acc Z) collected by the wristband device were first denoised and smoothed through the following two steps:

\begin{enumerate}
    \item \textbf{Calculate the Total Acceleration} \\
    The total acceleration (denoted as Total Acc) for each athlete at each data sampling point was computed based on the three-axis acceleration values recorded by the wristband sensor. The formula is:
    \begin{equation}
        \text{Total Acc} = \sqrt{\text{Acc X}^2 + \text{Acc Y}^2 + \text{Acc Z}^2}
    \end{equation}
    This step converts the three-dimensional acceleration data into scalar values that describe the overall motion state of the athlete at a given time point, facilitating subsequent data labeling, feature extraction, and motion recognition.

    \item \textbf{Differentiation and Smoothing Filter} \\
    A differentiation operation was performed on the Total Acc signal to obtain the differential acceleration (denoted as Diff Acc). A moving average filter with a window size of 20 was then applied to Diff Acc to compute the smoothed differential acceleration (denoted as SmoothDiff Acc), which enhanced motion boundary features and reduced noise.
\end{enumerate}

\subsubsection{PPG Signal Processing}

The raw PPG signals underwent a multi-step preprocessing pipeline. First, a third-order Butterworth bandpass filter (0.6--10 Hz) was applied to remove noise and retain relevant physiological frequency components. Next, a custom-designed peak detection algorithm was used to extract pulse peaks and compute RR intervals. Outlier RR intervals were manually identified and corrected using a jackknifing reconstruction method, resulting in a clean, continuous RR interval series for subsequent feature extraction.

\subsubsection{Data Labeling}

\begin{figure*}[h]
    \centering
    \includegraphics[width=0.6\textwidth, height=0.5\textheight, keepaspectratio]{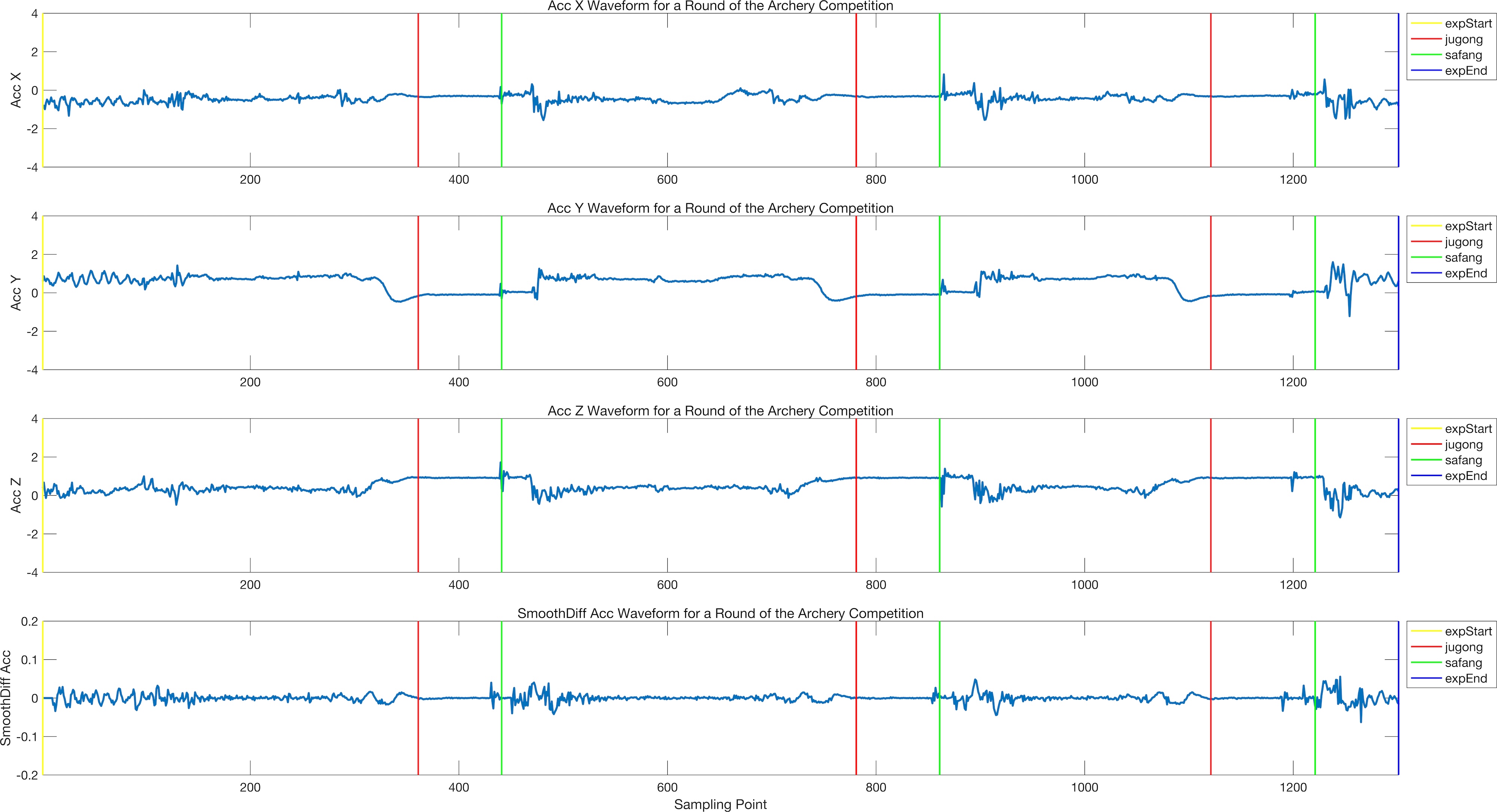}
    \caption{Example of accelerometer data waveforms with annotations}
    \label{fig:Labelled Figure}
\end{figure*}

Based on the SmoothDiff Acc and the manually annotated labels recorded during data collection (i.e., \texttt{Draw}, \texttt{Release}, \texttt{ExpStart}, \texttt{ExpEnd}), the archery actions were manually refined and calibrated following these steps:

\begin{enumerate}
    \item \textbf{Waveform Plotting and Label Display} \\
    Using the \texttt{Draw} label from the raw data annotations, the waveform of SmoothDiff Acc for 150 data points before and 300 data points after the \texttt{Draw} label was plotted. Additionally, the manually annotated labels within this data range were displayed. Based on the waveform and annotated labels, the archery actions corresponding to each arrow were manually annotated. This process continued until all \texttt{Draw} labels were manually verified. Notably, a complete draw-release action typically contains both a \texttt{Draw} and a \texttt{Release} label.

    An illustrative example is provided in Figure~\ref{fig:Labelled Figure}, which shows how these multi-dimensional acceleration signals, combined with manual labels, were jointly analyzed to refine the precise start and end points of each archery motion phase. This visual inspection allowed for validation and correction of the raw annotations, ensuring high-fidelity labeling for supervised learning.

    \item \textbf{Manual Annotation of Complete Archery Actions} \\
    Using the SmoothDiff Acc waveform and the annotated labels, complete draw-release actions were manually identified and annotated. Four mouse clicks were used for each action: (1) start of the draw phase, (2) end of the draw phase/start of the aim phase, (3) start of the release phase/end of the aim phase, and (4) end of the release phase (defined as the first peak after the release begins). Data between the start and end of the draw action were labeled as draw(\textit{jugong}), data for the aiming action were labeled as \texttt{AIM}, and data for the release action were labeled as release(\textit{safang}). These labels were stored in a newly created \texttt{labels} variable for supervised learning in the motion recognition task.
\end{enumerate}

\subsection{Modeling Tasks}


This section presents the modeling components for two tasks: motion phase recognition based on accelerometer data, and psychological stress classification based on physiological signals. Each task involves specific feature engineering and modeling strategies, detailed below.

\subsubsection{Motion Recognition}



During the data annotation process, we observed a distinct pattern in the \textit{ Draw-to-Release} phase. Specifically, during the transition from the \textit{Draw} to the \textit{Aim} phase, the accelerometer signals along the X-axis (Acc X) and Y-axis (Acc Y) initially increase and then decrease. During the \textit{Aim} phase, the signals from all three axes (Acc X, Acc Y and Acc Z) stabilize. Finally, in the \textit{Release} phase, the signals show an upward trend in the Acc X and Acc Z axes, while Acc Y shows a different pattern. This distinct signal behavior suggests that the task is feasible for modeling.

However, two challenges arise during the data collection process. First, ensuring the accuracy of the annotated labels is challenging due to delays introduced by manual annotation and data transmission from the recording tablet. Second, during the \textit{Aim} phase, the signal patterns can be unstable, despite the expectation of stability under ideal conditions. To address these challenges, we incorporated additional data sources to provide complementary information, enhancing the robustness of the model.

\paragraph{Feature Engineering}

To address these challenges, we incorporated two additional features to enhance the analysis. The first feature, Acceleration Total Load (total acceleration/deceleration; m/s²), represents the cumulative sum of all acceleration events over a defined analysis period\cite{totalAcc}. This metric captures the overall load derived from absolute acceleration and serves as a valuable tool in athlete monitoring. It can be used as an independent metric or as a supplementary variable to enhance the insights provided by threshold-based acceleration metrics\cite{totalAcc}. The formula is shown as follows:

\begin{equation}
\text{Total Acc} = \sqrt{\text{Acc X}^2 +\text{Acc Y}^2 +\text{Acc Z}^2}    
\end{equation}
To capture dynamic variations in acceleration during the analyzed period, we derived a smoothed differential feature from the three-axis accelerometer data (Acc X, Acc Y and Acc Z). The computation of this feature involved the following steps:
\begin{enumerate}

    \item Calculate the Total Acceleration

Compute the Euclidean norm of the acceleration components to obtain the overall magnitude of acceleration at each time step, denoted as \textbf{Total Acc}.This step provides a scalar representation of the overall acceleration intensity, independent of direction.
\item Temporal Differencing

To characterize the rate of change in acceleration magnitude, we apply a first-order temporal differencing operation to \textbf{Total Acc}, yielding
\begin{equation}
    \text{Diff}[t] = \text{Total Acc}[t] - \text{Total Acc}[t-1]
\end{equation}
Missing values introduced at the first time step are filled with 0 to maintain continuity in the time series. This feature effectively describes the magnitude of motion changes between consecutive intervals.
\item Noise Reduction via Rolling Mean Smoothing
To mitigate noise and emphasize meaningful trends in the data, the differenced signal(\textit{Diff}) is smoothed using a rolling window average. This smoothing process reduces the impact of random fluctuations and high-frequency noise arising from sensor inaccuracies, environmental disturbances, or minor, irrelevant movements. The smoothed differential is computed as:
\begin{equation}
\text{SmoothDiff}[t] = \frac{1}{N}\sum_{i=t-N+1}^{t}\text{Diff}[i]
\end{equation}
where $N$ is the window size. By averaging over a fixed number of consecutive data points, this approach ensures that the resulting feature captures the broader trend of motion changes, rather than transient spikes.

For boundary conditions (e.g., the start of the time series), missing values were replaced with 0 to preserve the continuity of the time series. This smoothing technique is particularly effective for improving signal stability and ensuring that the extracted feature reliably reflects the overall pattern of dynamic transitions in motion, making it more robust for downstream analysis.
\end{enumerate}

\paragraph{LSTM-Based Modeling}



Motion recognition from accelerometer signals requires capturing temporal dependencies and subtle transitions in movement. To this end, we employ a Long Short-Term Memory (LSTM) network, which is well-suited for modeling time series due to its memory cell structure and gating mechanisms that control information flow over time.

The input to the model includes raw accelerometer signals (\textit{Acc X, Acc Y, Acc Z}) and two derived features: \textit{Total Acc} and \textit{Smoothed Diff}. These features together capture both instantaneous motion and longer-term dynamics.

The LSTM model uses the \textit{tanh} activation function in the hidden layers to enable nonlinear transformations and dynamic memory updates. The output layer applies a \textit{sigmoid} activation to generate probabilistic predictions for binary classification, enabling the detection of key motion phases such as draw and release. The model’s ability to retain long-term dependencies makes it effective at identifying subtle transitions in sequential acceleration data, thus improving phase classification accuracy.


\paragraph{Prediction Strategy}

To identify the draw and release phases, the trained LSTM model outputs a sequence of probabilistic predictions for each sliding window of accelerometer data. A threshold is applied to these probabilities (\textit{e.g., 0.9}) to determine binary labels, where values above the threshold correspond to positive predictions.

To extract the indices corresponding to continuous draw and release phases, we use the following logic:
\begin{enumerate}
    \item Sliding Window Segmentation: The accelerometer data is divided into overlapping segments with a predefined window size and step length. Each segment is processed through the LSTM model to produce a prediction probability.
    \item Binary Label Assignment: Prediction probabilities are converted into binary labels based on the threshold. Labels greater than the threshold are classified as part of the draw phase.
    \item Consecutive Phase Detection: A post-processing step identifies consecutive positive labels as a single draw or release phase. The start and end indices of these segments are recorded for further analysis.
    \item Validation of Phase Durations: Segments are filtered based on their duration, ensuring they meet predefined minimum and maximum thresholds. Valid segments are retained as detected draw and release phases.
\end{enumerate}
 
The final output consists of the start and end indices of each detected phase, which can be mapped back to the original time series for temporal alignment and further evaluation. This approach ensures accurate detection of key motion phases, enabling robust motion recognition and supporting downstream tasks such as stress analysis.

\subsubsection{Stress Detection}

Psychological stress manifests through changes in autonomic nervous system activity, which can be monitored using heart rate variability (HRV) derived from photoplethysmogram (PPG) signals. We aim to classify stress levels based on physiological features extracted from the PPG time series.

\paragraph{Feature Engineering}


We utilize the Photoplethysmogram (PPG) signal to extract a range of basic variance-related features, which are essential for understanding physiological dynamics. These features, along with their mathematical formulations, are described as follows:
\begin{enumerate}
    \item Hear Rate(HR)

The number of heartbeats per minute, which provides a fundamental measure of cardiovascular activity and is sensitive to changes in physical or emotional states. The calculate method is 
\begin{equation*}
    \text{HR}=\frac{60}{\overline{\mathrm{NN}}}
\end{equation*}
\item SDNN (Standard Deviation of NN Intervals)

This reflects the variability of the inter-beat intervals (NN intervals). It is a global index of heart rate variability (HRV) and indicates the overall balance between sympathetic and parasympathetic nervous system activity. The SDNN is computed as:
\begin{equation}
\text{S D N N}=\sqrt{\frac{1}{N-1} \sum_{i=1}^N\left(\mathrm{NN}_i-\overline{\mathrm{NN}}\right)^2}    
\end{equation}

\item RMSSD (Root Mean Square of Successive Differences)

This measures short-term variability in heart rate by calculating the root mean square of successive differences between NN intervals. It is closely associated with parasympathetic activity, and calculated as:
\begin{equation}
    \text{R M S S D}=\sqrt{\frac{1}{N-1} \sum_{i=1}^{N-1}\left(\mathrm{NN}_{i+1}-\mathrm{NN}_i\right)^2}
\end{equation}

\item pNN20 and pNN50

These are the proportions of successive NN intervals that differ by more than 20ms (pNN20) or 50ms (pNN50), providing insights into short-term HRV and parasympathetic modulation. The formula is expressed as:
\begin{equation}
    \text{p N N x}=\frac{\text { Number of intervals where }\left|\mathrm{NN}_{i+1}-\mathrm{NN}_i\right|>x}{N-1} \times 100 \%
\end{equation}
where \textit{x} is  20 ms for pNN20 and 50 ms for pNN50.
\item HF (High-Frequency Power)

This component of HRV is derived from the power spectral density of the PPG signal, corresponding to the respiratory frequency range. It reflects parasympathetic nervous system activity.The power within the high-frequency range (typically 0.15–0.4 Hz) of the power spectral density (PSD) of the PPG signal, calculated using Fast Fourier Transform (FFT).
\item TF (Total Frequency Power)

The total power across the entire frequency spectrum of the PPG signal, indicating the overall energy of HRV.
\begin{equation}
    T F=\int_0^{\infty} \operatorname{PSD}(f) d f
\end{equation}
\item POB (Percentage of Oscillatory Beats)

This feature quantifies the proportion of beats exhibiting oscillatory characteristics, which may correlate with stress or physical exertion. The proportion of beats exhibiting oscillatory characteristics, often derived from threshold-based analysis of frequency-domain features.
\item SD1 and SD2 (from Poincaré Plots):

These are derived from Poincaré plots, where \textit{SD1} reflects short-term HRV and \textit{SD2} represents long-term HRV. Together, they provide a visual and quantitative understanding of autonomic balance. These are computed as:
\begin{equation}
    \begin{gathered}
S D 1=\sqrt{\frac{1}{2} \operatorname{Var}\left(\mathrm{NN}_{i+1}-\mathrm{NN}_i\right)} \\
S D 2=\sqrt{2 \operatorname{Var}(\mathrm{NN})-S D 1^2}
\end{gathered}
\end{equation}
\item Sample Entropy (sampEn)

This measures the complexity or irregularity of the PPG signal, offering insights into the dynamical system properties of heart rate regulation, and is defined as :

\begin{equation}
    \operatorname{sampEn}(m, r, N)=-\ln \frac{\mathrm{A}(m+1)}{\mathrm{B}(m)}
\end{equation}
where, 
\begin{itemize}
    \item \textit{m}: Embedding dimension,
    \item \textit{r}: Tolerance for similarity,
    \item A(\textit{m}+1): Number of (\textit{m}+1)-length patterns matching within tolerance \textit{r},
    \item B(\textit{m}): Number of m\textit{m}-length patterns matching within tolerance \textit{r}.
\end{itemize}

\end{enumerate}

\paragraph{MLP-Based Modeling}

For stress detection, we use a Multi-Layer Perceptron (MLP) model, which is applied to photoplethysmography (PPG)-derived features. MLP is selected for its ability to model nonlinear relationships and complex feature interactions in structured physiological data. It enables robust classification performance while maintaining computational efficiency and interpretability.

The selected input features include time-domain metrics (e.g., HR, SDNN, RMSSD, pNN20, pNN50), frequency-domain metrics (e.g., HF, TF), and nonlinear dynamics (e.g., SD1, SD2, sampEn), capturing multiple dimensions of heart rate variability (HRV) relevant to psychological stress.

This combination of physiological features and structured modeling provides a practical and reliable solution for stress classification tasks in real-world scenarios.

\paragraph{Prediction Strategy}

To enable supervised stress classification, we binarize self-reported Likert-scale ratings (1–5) into two classes: low stress (1–3) and high stress (4–5). This transformation simplifies the prediction task while preserving the ordinal structure of the original ratings.

Each training instance corresponds to a fixed-length window of PPG signal, from which HRV features are extracted. These features serve as input to the MLP classifier, which outputs a binary label indicating the predicted stress level. The prediction process is repeated across multiple windows to construct a robust sequence-level assessment of stress during archery sessions.

\section{Experiments}
\subsection{Motion Recognition}
\subsubsection{Data Preparation}




Based on the annotated accelerometer data described in the previous section, we constructed a labeled dataset for motion recognition. 

We applied a sliding window approach to segment the full-length signal, using a step size of 20 and a window size of 80 (corresponding to 4 seconds at a 20 Hz sampling rate). The choice of a 4-second window was based on empirical observations that the vast majority of draw-to-release motions fall within this duration. 

For labeling, we assigned a label of 1 to windows that predominantly contained a complete draw-to-release segment. Specifically, we examined the proportion of annotated positive data points (i.e., within the draw-to-release segment) contained in each window. If more than 50\% of the points in a window fell within a positive segment, the window was labeled as 1; otherwise, it was labeled as 0. This strategy ensures that each positive sample contains the full dynamic motion of interest, while avoiding fragmented or ambiguous segments. 

Due to the relative scarcity of motion compared to idle periods, negative samples heavily outnumber positives. To mitigate class imbalance, we applied a combination of under-sampling for the majority class and over-sampling for the minority class during training.

\subsubsection{Training Procedure and Evaluation Metrics}

Based on observed motion patterns in the dataset, we adopted a 70:30 split to train and test the model, enabling effective evaluation of its generalization capability. This partitioning ensures that the model is trained on sufficient data while retaining a realistic test distribution for generalization assessment.

To comprehensively evaluate model performance, we employ both standard classification metrics and deployment-oriented metrics suited for real-world applications. First, we assess classification accuracy and prediction balance within the curated dataset to evaluate the model's internal consistency. Second, we evaluate the model's robustness in deployment-like scenarios using metrics designed for real-time performance. The metrics are defined as follows:

We evaluate the model using standard classification metrics to assess its ability to distinguish between motion and non-motion segments. Specifically, \textbf{accuracy} reflects the overall proportion of correct predictions, while the \textbf{F1 score} captures the balance between precision and recall, making it especially useful for imbalanced classification.

To assess how well the model performs in real-world conditions, we introduce two practical evaluation metrics. The first metric, \textbf{Prediction Quantity Deviation (PQD)}, measures how closely the number of predicted positive samples matches the ground truth:

\begin{equation}
    \mathrm{PQD}=1-\left(\frac{\left|P_{\text {pred }}-P_{\text {true }}\right|}{P_{\text {true }}}\right)
\end{equation}

where $P_{pred}$ represents the predicted count of positive samples, and $P_{true}$ denotes the true count. A PQD value closer to 1 indicates strong agreement between predicted and true quantities, minimizing both underestimation and overestimation.

The second metric, \textbf{Sample-Level Accuracy (SLA)}, quantifies how often individual predictions match their ground truth labels. Each prediction is scored as 1 if it matches the actual label, and 0 otherwise. SLA is calculated as:

\begin{equation}
    \text { SLA }=\frac{N_{\text {correct }}}{N_{\text {total }}}
\end{equation}

where $N_{correct}$ is the number of correctly predicted samples, and $N_{total}$ is the total number of samples. This metric offers an intuitive assessment of prediction reliability in deployment scenarios.

By combining these metrics, we ensure a comprehensive evaluation of the model's performance that captures both its theoretical classification quality and its practical deployment reliability.

\subsection{Stress Detection}
\subsubsection{Data Preparation}



The dataset comprises PPG signals recorded during archery sessions, each accompanied by corresponding annotations. Each session is labeled with a self-reported stress level ranging from 1 (no stress) to 5 (high stress). For analysis, we binarize the stress labels into two classes: low stress (levels 1–3) and high stress (levels 4–5).

The continuous PPG signals are segmented using a sliding window of 30 seconds in length. The windows are aligned with the shooting sequence—from the start of the draw phase to the release phase—ensuring that each sample reflects the physiological state during aiming. From each window, we extract time-domain features (e.g., HR, SDNN, RMSSD), frequency-domain features (e.g., HF, TF), and nonlinear features (e.g., SD1, SD2, sampEn) to capture a comprehensive snapshot of the athlete’s physiological dynamics.

\subsubsection{Training Procedure and Evaluation Metrics}


The dataset is primarily divided into training and testing sets using a 70:30 split, though other variations (e.g., 80:20 or 70:20:10) are also explored. In our experiment, we adopt the 70:30 split for training and testing. The model is trained on labeled feature sets derived from the PPG signal and evaluated using several standard metrics. Specifically, we report \textbf{Accuracy}, \textbf{Precision}, \textbf{Recall}, and \textbf{F1 Score}. Accuracy measures the overall proportion of correct predictions, while Precision and Recall evaluate the model’s sensitivity to positive cases. F1 Score balances Precision and Recall, offering a robust summary of the model’s performance in imbalanced settings.

\section{Results}
\subsection{Motion Recognition}
\begin{table}[h!]
\centering
\caption{Performance comparison of different architectures and feature combinations}
\begin{tabular}{|l|l|c|c|c|c|}
\hline
\textbf{Architecture} & \textbf{Feature Combination}                     & \textbf{Accuracy (\%↑)} & \textbf{F1-score (\%↑)} & \textbf{PQD (\%↑)}   & \textbf{SLA (\%↑)}   \\ \hline
LSTM                  & Acc X, Acc Y, Acc Z                              & 84.53                 & 83.45                 & 96.34         & 67.86         \\ \hline
LSTM                  & Acc X, Acc Y, Acc Z,                             & 95.99            & 95.95            & 96.95         & 75.00         \\
                      & Total Acc, Smoothed Diff                         &                   &                   &                &                \\ \hline
BiLSTM                & Acc X, Acc Y, Acc Z,                             & 96.80            & 96.76            & 95.12         & 75.00         \\
                      & Total Acc, Smoothed Diff                         &                   &                   &                &                \\ \hline
\end{tabular}
\label{tab:shooting_detection_performance_comparison}
\end{table}




Table \ref{tab:shooting_detection_performance_comparison} compares the performance of different network architectures and feature combinations for detecting archery motion phases.

When using only the three-axis accelerometer data (Acc X, Acc Y, Acc Z), the baseline LSTM achieved an accuracy of 84.53\% and F1-score of 83.45\%. Incorporating the proposed Total Acceleration and Smoothed Differential Acceleration (SmoothDiff) features markedly improved performance, yielding 95.99\% accuracy and 95.95\% F1-score, with a 7.1\% increase in sample-level accuracy (SLA).

This improvement highlights the effectiveness of SmoothDiff in emphasizing meaningful temporal transitions while suppressing high-frequency noise, which enhances the model’s sensitivity to subtle motion boundary changes.

Comparing LSTM and BiLSTM architectures, the BiLSTM achieved a slightly higher accuracy (96.80\%) but at the cost of increased computational complexity and less real-time suitability. Given that archery actions follow a strictly sequential temporal pattern without future context dependence, the uni-directional LSTM provides a better balance between accuracy and efficiency.

Overall, these results validate the proposed feature engineering strategy and demonstrate that integrating SmoothDiff significantly strengthens the motion recognition model’s robustness for real-world deployment.

\subsection{Stress Detection}

\begin{table}[h]
\centering
\caption{Performance comparison of different methods using PPG features}
\resizebox{\textwidth}{!}{%
\begin{tabular}{|l|c|c|c|c|c|}
\hline
\textbf{Method}                 & \textbf{Acc (\%↑)} & \textbf{Prec (\%↑)} & \textbf{Rec (\%↑)} & \textbf{F1 (\%↑)} & \textbf{Leave-out Acc (\%↑)} \\ \hline
Logistic Regression             & 82.00              & 86.66              & 65.00              & 74.28             & 75.90                        \\ \hline
Decision Tree                   & 80.00              & 70.83              & 85.00              & 77.27             & 68.67                        \\ \hline
Random Forest                   & 76.00              & 75.00              & 60.00              & 66.66             & 71.68                        \\ \hline
Support Vector Machines         & 80.00              & 100.00             & 50.00              & 66.66             & 72.29                        \\ \hline
K-Nearest Neighbors             & 68.00              & 62.50              & 50.00              & 55.55             & 74.09                        \\ \hline
Naive Bayes                     & 74.00              & 88.88              & 40.00              & 55.17             & 70.48                        \\ \hline
Gradient Boosting Classifier    & 78.00              & 71.42              & 75.00              & 73.17             & 71.68                        \\ \hline
AdaBoost Classifier             & 78.00              & 73.68              & 70.00              & 71.79             & 68.07                        \\ \hline
Multi-Layer Perceptron          & 80.00              & 75.00              & 75.00              & 75.00             & 77.11                        \\ \hline
Linear Discriminant Analysis    & 82.00              & 100.00             & 55.00              & 70.96             & 78.31                        \\ \hline
LGBM Classifier                 & 74.00              & 64.00              & 80.00              & 71.11             & 72.29                        \\ \hline
XGBoost                         & 74.00              & 65.21              & 75.00              & 69.77             & 71.08                        \\ \hline
\end{tabular}%
}
\label{tab:stress_result}
\end{table}

The performance of various classifiers on PPG-derived features is summarized in Table \ref{tab:stress_result}.

Logistic Regression and Linear Discriminant Analysis (LDA) achieved the highest classification accuracy (82\%), but their relatively lower F1-scores (74.28\% and 70.96\%, respectively) suggest sensitivity to class imbalance. In contrast, the Multi-Layer Perceptron (MLP) achieved an F1-score of 75.00\% with a balanced trade-off between precision and recall, demonstrating stronger generalization in leave-out testing (77.11\% accuracy).

The MLP’s nonlinear feature integration capability likely contributes to its stable performance across both training and unseen data, confirming its suitability for physiological stress detection.

In summary, the results indicate that the proposed dual-model system—LSTM for motion recognition and MLP for stress detection—achieves high accuracy and robustness, providing a practical foundation for integrated biomechanical–physiological monitoring in archery training.

\section{Discussion and Future work}
    
    



The findings of this study demonstrate that wearable sensor data can be effectively leveraged for both motion phase recognition and stress level classification in archery. The introduction of the Smoothed Differential Acceleration feature notably enhanced the LSTM model’s ability to capture temporal transitions between phases, confirming the importance of refined kinematic features for precise movement recognition. Meanwhile, the use of HRV-based features in conjunction with an MLP classifier enabled reliable detection of psychological stress, suggesting a strong coupling between physiological signals and mental state during high-precision tasks.

Nevertheless, several limitations remain. The current dataset was collected from a relatively small cohort of adolescent archers under controlled conditions, which may limit generalizability to other age groups or competition settings. Furthermore, the stress labeling relied on self-reported measures, which could introduce subjective bias. Expanding the dataset and incorporating objective stress biomarkers (e.g., cortisol, galvanic skin response) would strengthen future analyses.

Future research will focus on three directions: (1) integrating rhythm control and inter-shot consistency metrics to better capture the dynamic stability of performance; (2) incorporating vision-based target scoring systems to link biomechanical and performance outcomes; and (3) exploring multimodal data fusion for real-time feedback and personalized training recommendations. These enhancements will transform the proposed system into a comprehensive, intelligent training platform for precision sports.

\section{Conclusion}

This study presents a novel wearable-based framework for integrated motion and stress analysis in archery. By combining accelerometer-derived Smoothed Differential Acceleration features with PPG-based HRV metrics, the system achieves high accuracy in both motion recognition (96.80\%) and stress classification (80\%). These results validate the feasibility of using low-cost, non-intrusive wrist-worn devices for comprehensive athlete monitoring in natural training environments.

The dual-model approach—LSTM for motion recognition and MLP for stress detection—provides an interpretable and scalable foundation for real-world implementation. Beyond its immediate application in archery, the proposed methodology can be extended to other precision sports requiring fine motor control and psychological resilience. Future work will focus on expanding datasets, enhancing multimodal integration, and deploying real-time adaptive feedback to further improve performance optimization and stress management.


\bibliographystyle{unsrt}  
\bibliography{references}

\end{document}